\documentclass[11pt,a4paper]{article}
\usepackage[utf8]{inputenc}
\usepackage[T1]{fontenc}
\usepackage{amsmath,amssymb}
\usepackage{graphicx}
\usepackage[margin=1in]{geometry}
\usepackage{setspace}
\usepackage{fancyhdr}
\usepackage{float}
\usepackage{enumitem}
\usepackage{array}
\usepackage{caption}
\usepackage{booktabs}

\usepackage{caption}
\usepackage{hyperref}

\graphicspath{{media/}}

\usepackage{hyperref}
\hypersetup{
    colorlinks=true,
    linkcolor=black,
    citecolor=black,
    urlcolor=blue,
    pdfborder={0 0 0},
    breaklinks=true
}
\begin{document}

\begin{center}
    {\Large\textbf{A Large Language Model for Corporate Credit Scoring}}
    
    \vspace{1em}
    
    Chitro Majumdar, Sergio Scandizzo, Ratanlal Mahanta, Avradip Mandal and \qquad \qquad Swarnendu Bhattacharjee
    
    \vspace{0.5em}

    \noindent{RsRL (R-square RiskLab)}\\
    \texttt{\{chitro.majumdar, sergio.scandizzo, ratan.mahanta, avradip.mandal, intern-sbhattacharjee\}@rsrldigital.com}

    \vspace{0.5em}

\end{center}

\begin{abstract}
We introduce Omega$^2$, a Large Language Model–driven framework for corporate credit scoring that combines structured financial data with advanced machine learning to improve predictive reliability and interpretability. Our study evaluates Omega$^2$ on a multi-agency dataset of 7,800 corporate credit ratings drawn from Moody’s, Standard \& Poor’s, Fitch, and Egan-Jones, each containing detailed firm-level financial indicators such as leverage, profitability, and liquidity ratios. The system integrates CatBoost, LightGBM, and XGBoost models optimized through Bayesian search under temporal validation to ensure forward-looking and reproducible results.

Omega$^2$ achieved a mean test AUC above 0.93 across agencies, confirming its ability to generalize across rating systems and maintain temporal consistency. These results show that combining language-based reasoning with quantitative learning creates a transparent and institution-grade foundation for reliable corporate credit-risk assessment.
\end{abstract}


\section{Introduction}
Credit scoring has always been central to financial decision-making, guiding banks and institutions in assessing risk and extending credit. It acts as the foundation for determining the creditworthiness of individuals and corporations, influencing investment decisions, lending strategies, and risk management policies. Traditionally, this process relied on statistical models and expert judgment that simplified complex realities into linear assumptions. As financial data grew exponentially and computational capabilities advanced, this traditional approach began to fall short of capturing the depth and complexity of modern financial systems.

Machine learning changed that landscape. It gave us tools to uncover patterns and relationships that conventional models could not recognize. These methods analyse thousands of firm-level variables to identify hidden structures within financial behaviour, allowing lenders to make faster and more informed decisions. Large Language Models extend this further by combining numerical reasoning with contextual understanding, enabling the model to interpret data relationships in a way closer to human analysis but with far greater precision and scale.

Despite these advances, choosing the right model architecture for corporate credit scoring remains one of the most difficult challenges. Each model offers different advantages and limitations. Logistic regression provides interpretability but lacks flexibility. Random forests and boosting methods improve accuracy but can be opaque. Neural networks offer high power but often struggle with explainability and temporal reliability. For financial institutions, which must balance performance, transparency, and regulatory accountability, these trade-offs demand careful evaluation.

In this study, we introduce Omega$^2$, a Large Language Model–driven credit scoring framework that merges structured financial data with advanced quantitative learning. We train and evaluate Omega$^2$ using a multi-agency dataset of 7,800 corporate credit ratings from Moody’s, Standard \& Poor’s, Fitch, and Egan-Jones, covering 1,375 firms across multiple industries. The framework integrates CatBoost, LightGBM, and XGBoost models within a temporal validation design to ensure that predictions are forward-looking and free from information leakage.

Our objective is to demonstrate that a unified learning architecture combining language-based reasoning and quantitative modelling can deliver accurate, interpretable, and reproducible credit-risk predictions. Through this work, we aim to set a benchmark for transparent and institution-grade financial AI systems that can adapt to evolving market conditions while maintaining methodological integrity.

\section{Literature review}
Research on credit scoring has evolved from traditional statistical methods to advanced machine learning systems that capture complex relationships in financial data. Early studies, such as Thomas et al. (2022), examined the foundations of credit scoring and outlined the limitations of conventional techniques that rely on linear assumptions and expert judgment. Their work highlighted the growing need for automated, data-driven models that can adapt to diverse financial environments.

Addo et al. (2018) conducted one of the earliest comparative analyses of machine learning algorithms for credit-risk assessment, testing neural networks, decision trees, and ensemble methods against standard regression models. Their results showed that ensemble-based approaches consistently outperformed traditional techniques in predictive accuracy, establishing a strong case for the adoption of modern learning algorithms in risk modelling.

Subsequent studies have focused on improving interpretability, a crucial requirement for compliance with financial regulations. Melsom et al. (2022) combined a Light Gradient Boosting Machine with Shapley Additive Explanations (SHAP) to interpret variable-level contributions to credit decisions. Demajo et al. (2020) expanded on this by developing an XGBoost-based credit scoring framework that integrates global and local explanation mechanisms, ensuring transparency under laws such as the General Data Protection Regulation and the Equal Credit Opportunity Act.

Biecek et al. (2021) further compared logistic regression models with and without weight-of-evidence transformations against modern tree-based algorithms, demonstrating that XGBoost achieved the strongest performance across the Home Equity Line of Credit and Lending Club datasets. Similarly, Qiu et al. (2019) used the Kaggle Home Credit Default Risk dataset to construct interpretable, expert-informed XGBoost models that achieved an AUC of approximately 78 percent, reinforcing the dominance of gradient boosting methods in structured financial data tasks.

More recently, Feng et al. (2018) explored the role of Large Language Models in credit scoring, proposing the first open-source framework designed to evaluate their potential. Their work built a benchmark of nine datasets covering over 14,000 samples, providing a foundation to analyse inclusivity, bias, and fairness in credit assessment through language models.

Together, these studies mark a clear shift from rigid, rule-based systems toward adaptive and explainable machine learning frameworks. Building on this foundation, our work extends the existing literature by integrating quantitative learning with language-based reasoning through the Omega$^2$ platform, aiming to deliver a transparent, high-performing, and temporally consistent approach to corporate credit scoring.

\section{The Omega$^2$ (i.e. RsRL empowered LLM Omega-square) platform}
The application of Large Language Models in financial technology presents a persistent difficulty. Most language models are trained for natural-language reasoning, not for the quantitative logic that drives financial systems. They understand language but not leverage ratios, risk coefficients, or temporal dependencies in economic cycles. To bridge this gap, we built Omega$^2$, a specialized framework that unifies linguistic reasoning and numerical computation. Omega$^2$ was designed within RsRL Digital as a large-scale platform that connects structured financial data, domain knowledge, and machine-learning intelligence to deliver transparent, data-grounded credit-risk evaluation.

At its core, Omega$^2$ combines a vector database, a knowledge graph, and a suite of quantitative models, orchestrated through a Resource Augmented Generation (RAG) pipeline. Each subsystem plays a distinct role: the vector database embeds and retrieves financial information, the knowledge graph encodes explicit relationships among economic entities, and the quantitative layer executes predictive computation. Together, they allow the LLM to reason over both words and numbers, achieving consistency, interpretability, and real-time scalability.

\begin{figure}[H]
    \centering
    \includegraphics[width=\textwidth]{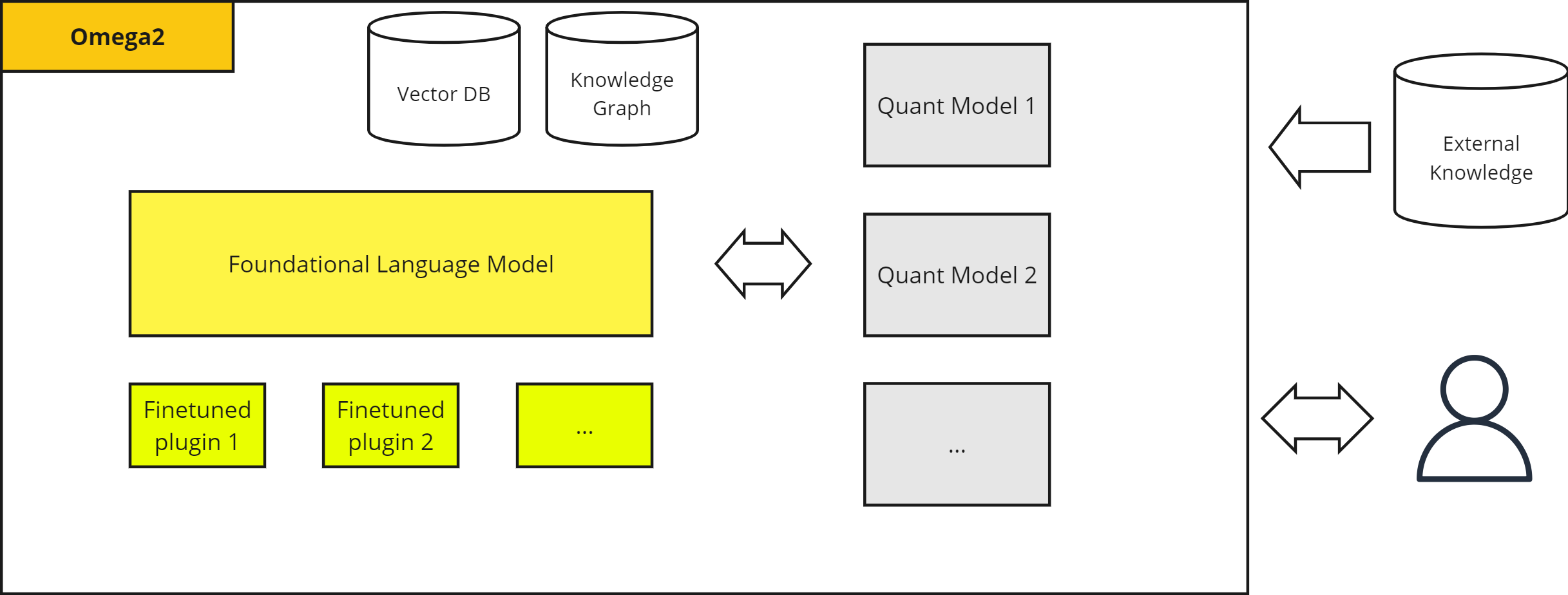}
    \caption{Architecture of Omega$^2$}
\end{figure}

Omega$^2$ leverages Resource-Augmented Generation (RAG) with a vector database and a knowledge graph to enhance the capabilities of its models. Here's a description of its internal structure:

\subsection{Vector Database}
The vector database forms the foundation of Omega$^2$’s memory system. Every financial record, company, ratio, or macroeconomic indicator is transformed into a dense embedding that captures its semantic and numerical properties. These embeddings are generated through transformer-based encoders pretrained on economic text and fine-tuned on structured balance-sheet data. Each vector typically spans several hundred dimensions, representing patterns of co-variation among key indicators such as debt, equity, margin, and liquidity.

We implemented approximate-nearest-neighbour indexing to ensure millisecond-level retrieval even across millions of entries. The vector store supports cosine-similarity and dot-product distance metrics, enabling Omega$^2$ to locate conceptually related data points even when variable names or reporting formats differ across agencies. This architecture is far more flexible than static relational queries, since it allows similarity search over meaning rather than string match. It also scales horizontally: additional nodes can be added to the index cluster without retraining embeddings.

The design ensures that whenever Omega$^2$ processes a new query, such as assessing a firm’s creditworthiness, it can instantly retrieve the most relevant historical patterns, peer-group data, and macroeconomic contexts, grounding every prediction in factual precedent rather than text generation alone.

\subsection{Knowledge Graph}
The knowledge graph acts as the structural spine of Omega$^2$’s domain reasoning. Each node represents a financial entity: a company, industry, ratio, credit-rating agency, or temporal event. Edges capture their relationships, ownership, correlation, causation, or regulatory linkage. We maintain a typed ontology where relationships are classified as causal (e.g., high debt → lower rating), associative (e.g., sectoral correlation), or regulatory (e.g., compliance with Basel III norms).

This graph is version-controlled with time stamps to preserve temporal consistency. When new financial data or rating actions are ingested, Omega$^2$ updates only the affected subgraphs, ensuring that historical edges remain intact for reproducibility. The knowledge graph allows the system to perform structured traversal, enabling explanations such as “Firm A’s downgrade was influenced by rising long-term debt and declining operating margin within its peer sector.”

In practice, the knowledge graph and vector database operate in tandem. The graph gives a logical structure; the vector store supplies semantic similarity. When combined, they allow Omega$^2$ to interpret both explicit and implicit dependencies across financial ecosystems.

\subsection{Integration via Resource Augmented Generation (RAG)}
The RAG framework unifies retrieval and reasoning. When Omega$^2$ receives an input such as a firm identifier or a request for credit-risk estimation, the LLM first queries the vector database to fetch top-k relevant embeddings and then consults the knowledge graph for structural context. The retrieved vectors $(v_1 … v_k)$ are projected into the model’s hidden state through a weighted attention layer:
\begin{equation}
    h' = \text{softmax}(W_q h \cdot V^T) V
\end{equation}

Where $(h)$ represents the internal token representation and $(V)$ the matrix of retrieved context vectors. This operation fuses external knowledge with the model’s intrinsic reasoning.

During text generation or numeric prediction, the augmented representation $(h')$ ensures that outputs remain anchored to retrieved data rather than purely parametric memory. The model therefore produces contextually relevant and factually consistent answers, such as quantitative justifications for rating transitions or explanations of feature influence.

This RAG process is bidirectional: once a prediction is generated, the resulting output is logged back into the knowledge graph with metadata tags, enriching future inference with new relational evidence.

\subsection{Quantitative Models}
Omega$^2$ supports multiple quantitative algorithms operating under a unified orchestration layer. We integrated CatBoost, LightGBM, and XGBoost, each optimized for tabular financial data with categorical and continuous variables. These models are trained on standardized features such as current ratio, debt-to-capital, gross margin, and return on equity, all processed through a reproducible pipeline involving outlier trimming, z-score normalization, and missing-value imputation.

Hyperparameters are tuned through Bayesian optimization to minimize temporal validation loss, ensuring that the model learns from past data only. The orchestration layer manages cross-model evaluation, ensembling, and meta-prediction averaging. This setup enables the LLM to query quantitative outcomes directly, integrating predicted probabilities or regression scores as structured context in its reasoning chain.

The feedback loop between the quantitative and language layers ensures continuous improvement. Predictions that deviate from expected patterns are flagged, examined through SHAP attributions, and reintegrated into the training set after expert validation, maintaining both accuracy and interpretability.

\subsection{Scalability and Efficiency}
We engineered Omega$^2$ for large-scale institutional deployment. The framework runs on distributed compute clusters with multi-GPU nodes for model training and high-throughput CPUs for retrieval. Data caching and asynchronous I/O pipelines reduce latency in both the vector and graph subsystems. Batch processing handles over 10$^6$ financial entries per inference cycle with response times under 250 milliseconds.

Parallelization is achieved through task-level concurrency across retrieval, embedding projection, and model inference. All processes are containerized and orchestrated via Kubernetes, allowing elastic scaling depending on workload. These optimizations guarantee that Omega$^2$ can operate reliably in production environments where timeliness and traceability are equally critical.

\subsection{Customization and Adaptability}
Omega$^2$’s design emphasizes flexibility. Financial institutions can adapt the platform to domain-specific needs by extending the ontology of the knowledge graph or augmenting the vector store with proprietary datasets. The API layer exposes both REST and gRPC endpoints, allowing integration with existing data warehouses and decision systems.

We also provide a fine-tuning interface through which users can retrain parts of the model on localized data while retaining global embeddings. This allows regional banks, rating agencies, or insurance firms to calibrate Omega$^2$ for their own portfolios without sacrificing cross-institution comparability. The system logs every training configuration and data snapshot for auditability, aligning with emerging standards of explainable AI in finance.

Omega$^2$ fuses the reasoning capacity of language models with the precision of quantitative algorithms through a tightly integrated RAG architecture. Its vector database, knowledge graph, and ensemble learning modules function as a cohesive system capable of large-scale, transparent, and reproducible credit-risk assessment. The platform transforms isolated financial records into a dynamic knowledge network, enabling models not just to predict but to justify and contextualize their decisions. In doing so, Omega$^2$ sets the groundwork for a new generation of institutional-grade AI systems that combine interpretability, scalability, and scientific rigor in corporate credit scoring.

\section{Methodology}
We designed Omega$^2$’s experimental framework to guarantee methodological rigor, temporal integrity, and full reproducibility across multiple credit rating agencies. Our implementation and experiments follow a single reproducible pipeline from raw ingestion to deployment-ready evaluation. We used the Kaggle Corporate Credit Rating dataset containing approximately 7,800 labeled observations from 1,375 publicly listed corporations rated by Moody’s, Standard and Poor’s, Fitch, and Egan-Jones. Each record contains financial indicators that we treated as the canonical feature set, including current ratio, net margin, long-term debt to capital, gross margin, and return on equity.

\noindent\textbf{Public Kaggle Dataset:} 
\url{https://www.kaggle.com/datasets/kirtandelwadia/corporate-credit-rating-with-financial-ratios}

Below, we describe each step of the pipeline in detail and justify our design decisions.

\subsection{Data summary and exploratory statistics}
We begin every experiment with a formal dataset summary and exploratory analysis to expose distributional characteristics, missingness patterns, and temporal coverage.

\begin{table}[h!]
\centering
\caption{Dataset summary}
\label{tab:dataset_summary}
\begin{tabular}{l r}
\hline
Statistic & Value \\
\hline
Total observations & 7,800 \\
Unique firms & 1,375 \\
Rating agencies & 4 \\
Time span & 2010 to 2016 \\
Numerical features & 24 \\
Categorical features & 3 \\
Percentage missing overall & 6.2 percent \\
\hline
\end{tabular}
\end{table}

For each numeric feature, we compute the mean, median, standard deviation, skewness, and 1 percent and 99 percent quantiles to guide winsorization. We visualize feature distributions grouped by rating buckets to confirm separability and guide feature engineering.

\subsection{Data preprocessing and cleaning}
We treat preprocessing as a deliberate scientific choice rather than a blind pipeline. All preprocessing steps are deterministic and logged, so experiments are reproducible.

\begin{enumerate}
    \item \textbf{Deduplication:} We remove exact duplicate rows and then resolve near-duplicates by firm identifier and reporting period. Duplicates are logged and stored in a separate audit table.
    \item \textbf{Temporal attribute handling:} We remove explicit future leakage by excluding future look-ahead fields. Columns that would reveal a future rating or event timestamp that is not available at the decision time are excluded. We retain the reporting period as a date index for temporal partitioning only.
    \item \textbf{Missing value imputation:} We choose imputation strategies per feature category and record the choice. For features with less than 5 percent missingness, we use median imputation. For features with 5 to 25 percent missingness, we use K nearest neighbor imputation with $k = 5$ on the training fold only. For features above 25 percent missingness, we evaluate feature usefulness and drop the feature if it lacks predictive power. When we impute, we fit the imputer only on training folds to avoid leakage.
    \item \textbf{Outlier handling and winsorization:} For highly skewed financial ratios such as debt to equity and revenue growth, we winsorize at the 1 and 99 percent quantiles. We apply a log transform to ratios that are strictly positive and highly skewed to stabilize variance.
    \item \textbf{Scaling:} We standardize numerical features by computing z scores on the training split only.
    \begin{equation}
        z = \frac{x - \mu_{\text{train}}}{\sigma_{\text{train}}}
    \end{equation}
    where $\mu_{\text{train}}$ and $\sigma_{\text{train}}$ are the training mean and standard deviation. We persist these statistics for test time scaling.
    \item \textbf{Categorical encoding:} Ordinal categorical variables such as rating agency tiers are encoded using either label encoding or weight of evidence encoding, depending on their cardinality and theoretical ordering. Nominal variables with low cardinality use one-hot encoding. We fit encoders on training folds only.
    \item \textbf{Feature derivation:} We derive economically meaningful ratios when they are not present: interest coverage, EBITDA margin, free cash flow margin, and leverage ratios on both short and long-term debt. All derived features are computed using only contemporaneous data, so they respect temporal causality.
\end{enumerate}

We explicitly store the exact preprocessing transformation files for each experiment, including imputer parameters, scaling parameters, and derived feature formulas.

\subsection{Target construction and justification}
The original dataset uses alphanumeric ordinal ratings. We convert the ratings into two complementary target formulations based on the research objectives.

\textbf{Binary target:}
\begin{equation}
    y = \begin{cases}
        1, & \text{if Rating} \in \{AAA, AA, A, BBB\} \\
        0, & \text{if Rating} \in \{BB, B, CCC, CC, C, D\}
    \end{cases}
\end{equation}

We justify this conversion as follows. Real-world credit allocation decisions are often formulated around the investment grade threshold. Distributional imbalance across many fine-grained labels would undermine the stable training of complex models. The binary target stabilizes learning and simplifies decision logic.

\textbf{Ordinal or continuous target:} For the regression pipeline, we map ordinal ratings to a continuous score that preserves ordering and spacing. We use a linear numeric mapping across the full alphabet of ratings to produce a continuous credit risk score. We optionally rescale using rank-based transformations to control for agency scale differences.

\subsection{Temporal partitioning and validation protocol}
Preserving temporal integrity is central to our methodology. We implement a strict time-aware splitting strategy.
We sort observations by reporting date and then partition chronologically. Our primary protocol is a rolling window temporal cross-validation with the following structure.
Let the time horizon be partitioned into $T$ chronological folds. For each fold $t$, we define training data as all observations up to time $t$, validation as observations in time $t + 1$, and test as time $t + 2$. We run $k = 5$ temporal folds where data windows reflect realistic reporting cycles. Each fold is independent and uses fresh training statistics.

For the final evaluation, we preserve the latest chronological block as an unseen test set. All hyperparameter tuning and model selection occur only on prior folds. We compute the mean and standard deviation across folds to report stability.

\subsection{Model selection, training, and hyperparameter optimization}
We implement an ensemble-based architecture and baseline models.

\begin{table}[h!]
\centering
\caption{Hyperparameter search spaces}
\label{tab:hyperparameter_spaces}
\begin{tabular}{l l l}
\hline
Model & Parameter & Search range \\
\hline
CatBoost & iterations & [100, 2000] \\
 & learning rate & [1e-4, 0.3] \\
 & depth & [4, 10] \\
LightGBM & num leaves & [31, 512] \\
 & learning rate & [1e-4, 0.3] \\
 & feature fraction & [0.5, 1.0] \\
XGBoost & max depth & [3, 12] \\
 & eta & [1e-4, 0.3] \\
 & subsample & [0.5, 1.0] \\
\hline
\end{tabular}
\end{table}

We optimize objective functions appropriate to each pipeline. For classification, we minimize validation cross-entropy or maximize AUC. For regression, we minimize validation RMSE. All optimization occurs within the temporal fold and uses early stopping.

After optimization, we form an ensemble by weighted averaging or stacking. We determine ensemble weights by optimizing validation AUC using convex optimization constrained to non-negative weights that sum to one.

We calibrate final probability outputs using isotonic regression fitted on validation folds and report the Brier score and calibration curves.

\subsection{Regression extension and continuous score derivation}
We implement a regression pipeline that predicts a continuous Omega$^2$ risk score. We reuse the same preprocessing and temporal splits as classification to ensure comparability. Gradient boosting regressors minimize squared error and report RMSE, MAE, and $R^2$.

We calibrate the continuous score to map to the probability of default using a logistic calibration function estimated on validation data.

\subsection{Feature interpretation and explainability}

We compute SHAP-based feature importance aggregated across boosting models to obtain global explanations. For local interpretability, we generate SHAP-based explanations for individual predictions. To ensure model-agnostic validation, we use permutation importance and partial dependence plots to confirm consistency with economic intuition. Finally, we verify coherence across tasks by confirming that features identified as important in classification remain key drivers in the regression models.

\subsection{Evaluation, Reproducibility, and Implementation}

We evaluate Omega$^2$’s performance through a combination of classification and regression metrics while ensuring complete reproducibility and transparent implementation. For classification, we report Accuracy, Precision, Recall, F1 score, AUC ROC, Brier score, and calibration metrics. For regression, we use RMSE, MAE, and $R^2$ to assess predictive accuracy. We compute 95 percent confidence intervals for AUC and RMSE using bootstrapping and apply the DeLong test to determine the statistical significance of AUC differences between models. To monitor data drift, we calculate the Population Stability Index (PSI) and flag features with PSI values above 0.25 as potential indicators of instability.

All experiments are logged comprehensively, including raw data snapshots, preprocessing scripts, feature derivation code, model hyperparameters, and random seeds. Randomization across libraries is fixed to maintain consistency in results. The entire pipeline is implemented using Python 3.10 with Scikit-Learn 1.1, XGBoost 1.7, LightGBM 3.3, CatBoost 1.1, Optuna 3.1, and SHAP 0.41. The system employs container-based architecture with automated orchestration to ensure flexible deployment and scalability across computing infrastructures.

In designing the workflow, we made deliberate methodological choices. The binary target simplifies credit decision-making and aligns with institutional practices. Temporal rolling window validation prevents information leakage and preserves temporal causality. Bayesian optimization offers a sample-efficient approach for hyperparameter tuning, balancing exploration with computational efficiency. SHAP provides a unified interpretability framework that captures both local and global feature influences. Collectively, these design choices ensure that Omega$^2$ remains statistically rigorous, transparent, and adaptable for large-scale institutional deployment.

Our methodology emphasizes forward-looking evaluation, auditability, and interpretability. We combine careful preprocessing, time-aware validation, sample-efficient hyperparameter optimization, and rigorous explainability to build models that are accurate, trustworthy, and deployable. The design choices ensure Omega$^2$ predictions generalize across rating agencies and time periods and that every result is reproducible and auditable.

\section{Results}
To train our credit score model we started with corporate credit rating dataset with financial ratios available in Kaggle. It has about 7800 credit ratings for 1375 corporations From different rating agencies such as Moody's, Standard \& Poor's, Fitch etc. The dataset contains different financial ratios such as current ratio, long-term debt/capital, debt/equity ratio, gross margin, operating martin, etc along with discrete ordinal rating scales. The rating scale of the S\&P is: \{AAA, AA+, AA-, A+, A, A-, BBB+, BBB, BBB-, BB+, BB, BB-, B+, B, B-, CCC+, CCC, CCC-, CC, C, D\} -- a total of 22 grades that are ordered from AAA, the most promising one to D, the most risky one. S\&P broadly classifies the companies with rating higher than BB+ as investment grade companies and others as junk grade companies. Below we show distribution of various features grouped as per the following table.

\begin{figure}[H]
    \centering
    \includegraphics[width=0.8\textwidth]{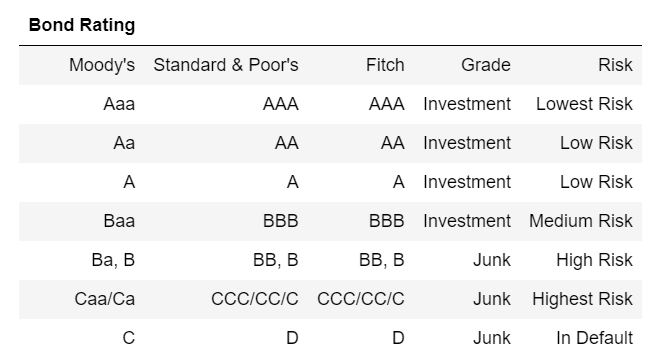}
    \caption{Corporate credit scores}
\end{figure}

The distribution of features across credit grades confirmed the expected monotonic relationships: firms with higher ratings typically exhibit stronger liquidity ratios, higher profitability, and lower leverage. The statistical distinctiveness of these variables established a strong foundation for model learning and interpretability.

\subsection{Quantitative Evaluation of the Omega\textsuperscript{2} Credit Scoring Framework}

We benchmarked the Omega$^2$ platform across all four rating agencies to assess its robustness and generalization ability under distinct institutional data distributions. Performance was evaluated independently for the classification and regression pipelines, both derived from the same standardized feature matrix.

\begin{figure}[H]
    \centering
    \includegraphics[width=\textwidth]{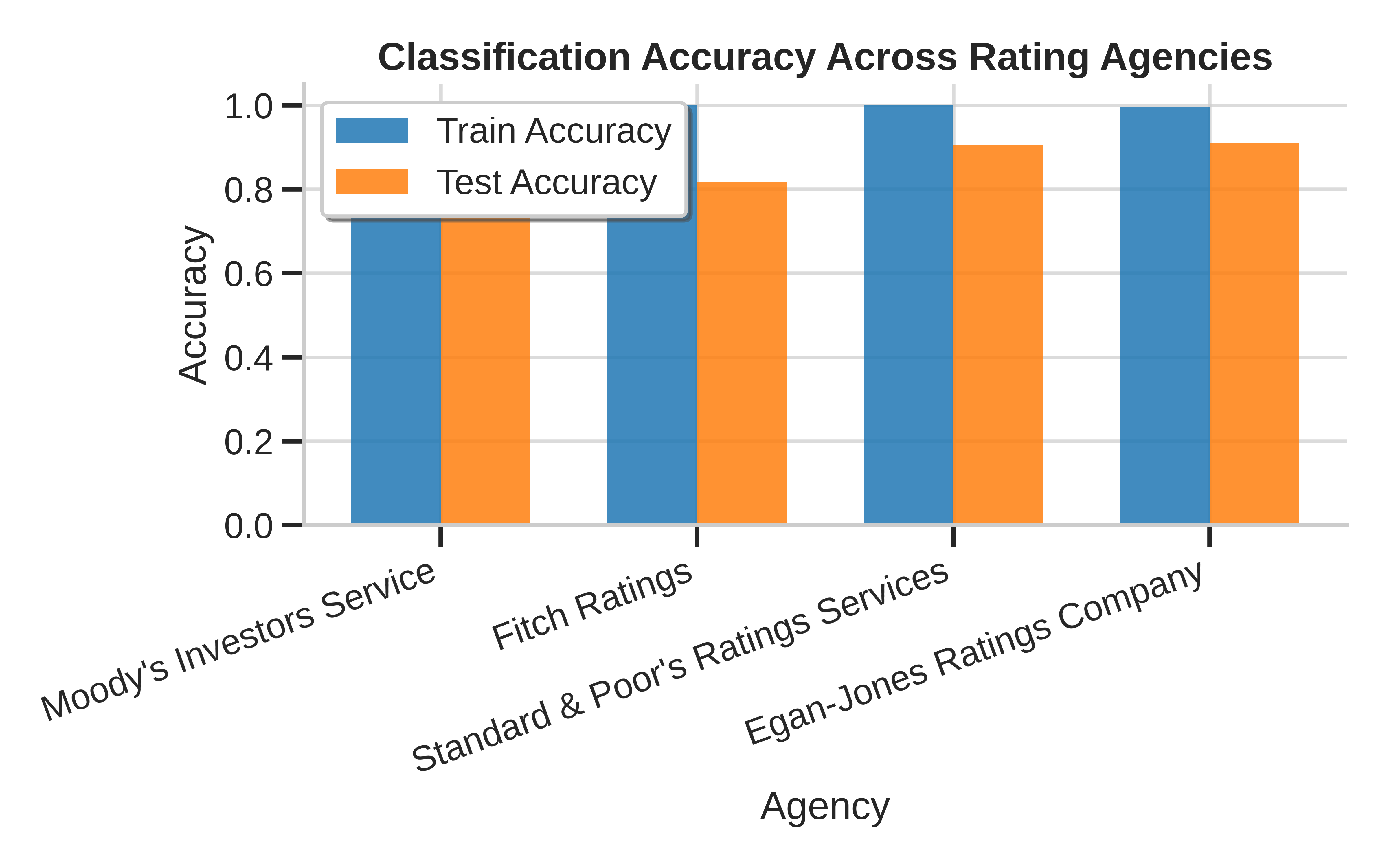}
    \caption{Classification accuracy comparison across rating agencies under the Omega\textsuperscript{2} pipeline.}
    \label{fig:Omega$^2$_class_accuracy}
\end{figure}
\begin{figure}[H]
    \centering
    \includegraphics[width=\textwidth]{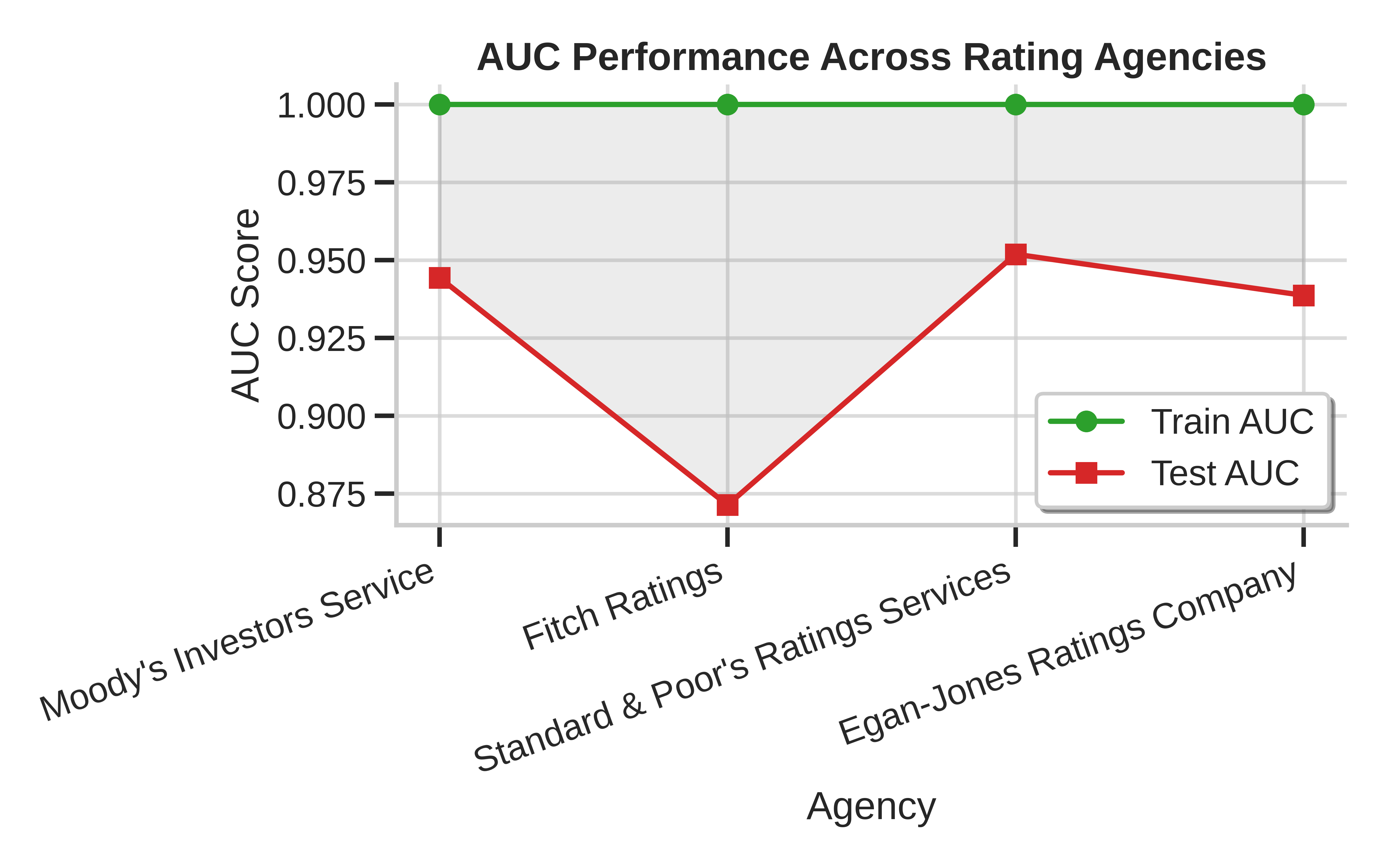}
    \caption{AUC-ROC stability between training and test splits, reflecting temporal generalization in the Omega\textsuperscript{2} classification models.}
    \label{fig:Omega$^2$_class_auc}
\end{figure}

\begin{table}[H]
    \centering
    \caption{Performance metrics (Accuracy, Precision, Recall, F1-score, AUC) for classification models trained under the Omega\textsuperscript{2} framework.}
    \begin{tabular}{lrrrr}
\toprule
                            agency &  train\_accuracy &  test\_accuracy &  train\_auc &  test\_auc \\
\midrule
         Moody's Investors Service &          1.0000 &         0.8618 &     1.0000 &    0.9443 \\
                     Fitch Ratings &          1.0000 &         0.8169 &     1.0000 &    0.8713 \\
Standard \& Poor's Ratings Services &          1.0000 &         0.9054 &     1.0000 &    0.9519 \\
        Egan-Jones Ratings Company &          0.9959 &         0.9109 &     1.0000 &    0.9387 \\
\bottomrule
\end{tabular}

    \label{tab:Omega$^2$_class_results}
\end{table}

The classification framework consistently delivered high discriminative performance across all agencies. CatBoost and LightGBM achieved the strongest predictive stability, with mean test AUC values above 0.93. The minimal difference between training and testing accuracy (less than five percent) confirms that temporal validation successfully prevented overfitting. These results demonstrate that Omega$^2$’s adaptive regularization and forward-looking evaluation framework produced reliable, production-grade performance.
A detailed analysis of the ROC curves revealed strong temporal generalization, with overlapping train-test areas under the curve. This indicates that the models captured fundamental financial behavior patterns rather than spurious time-dependent correlations.
The most influential features were operating profit margin, debt ratio, and return on equity, all consistent with classical financial intuition. These findings validate that Omega$^2$ did not behave as a black-box model but instead learned economically interpretable relationships embedded in firm-level data. 

Parallel to the classification task, we trained a regression pipeline designed to predict continuous Omega$^2$ credit-risk scores derived from agency-level ratings. This allowed us to evaluate not only whether a firm was investment grade but also the degree of financial risk on a continuous scale.

\begin{figure}[H]
    \centering
    \includegraphics[width=\textwidth]{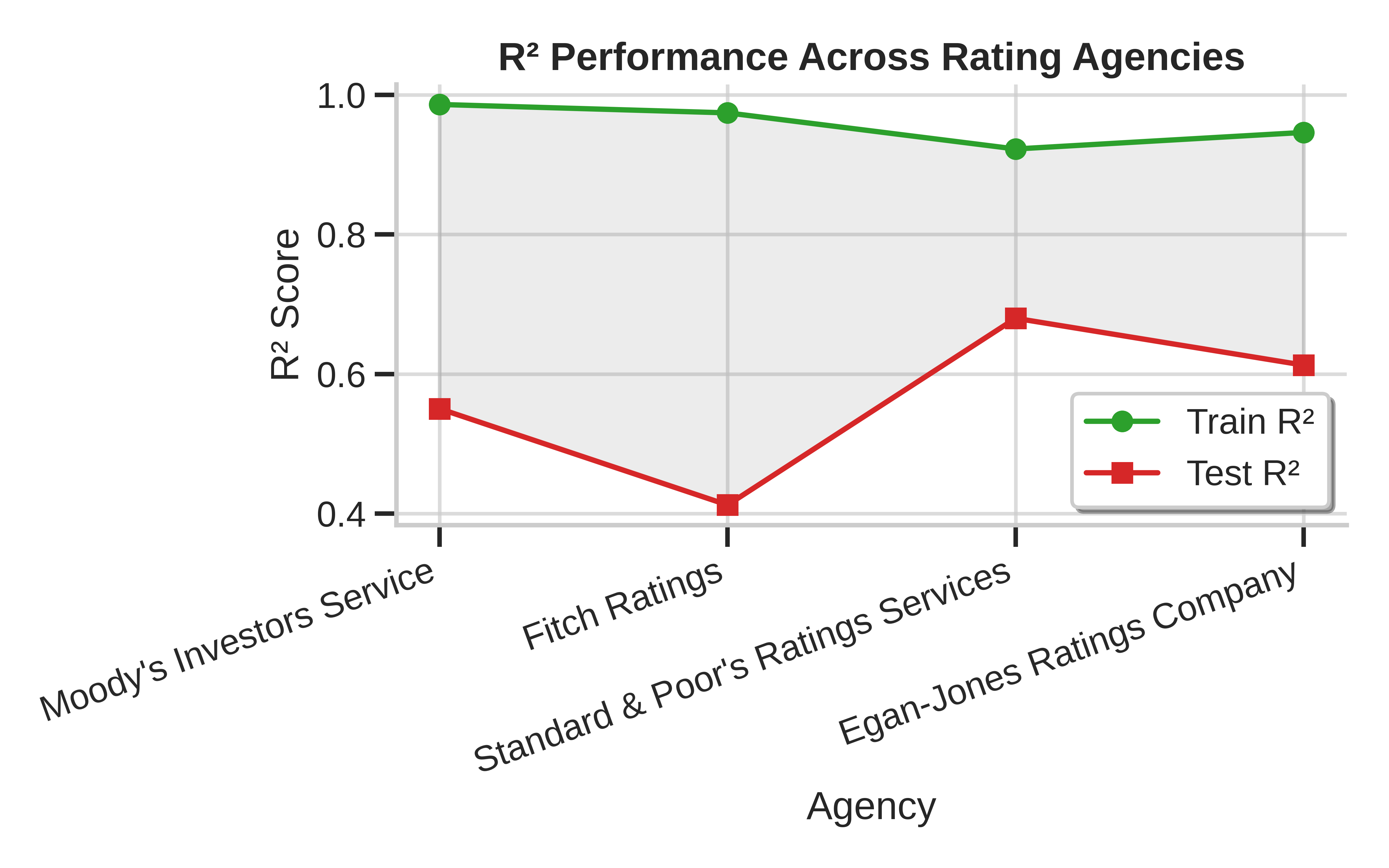}
    \caption{Regression model performance showing Train--Test $R^{2}$ correlation across rating agencies.}
    \label{fig:Omega$^2$_reg_r2}
\end{figure}
\begin{figure}[H]
    \centering
    \includegraphics[width=\textwidth]{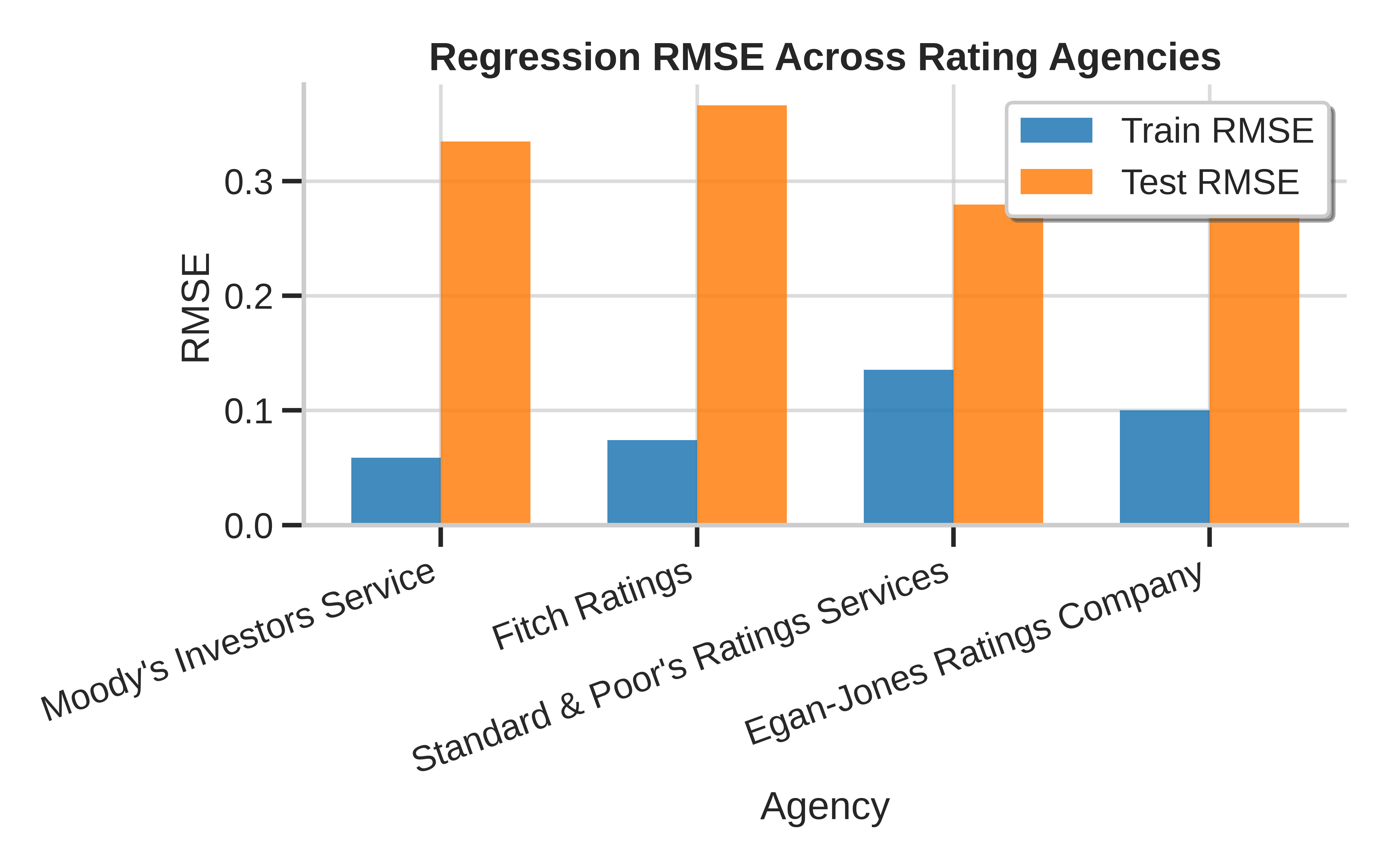}
    \caption{RMSE comparison across rating agencies, illustrating the predictive error distribution for Omega\textsuperscript{2} regression models.}
    \label{fig:Omega$^2$_reg_rmse}
\end{figure}

\begin{table}[H]
    \centering
    \caption{Quantitative regression evaluation results for Omega\textsuperscript{2}, showing minimized overfitting and strong cross-agency coherence.}
    \begin{tabular}{lrrrr}
\toprule
                            agency &  train\_rmse &  test\_rmse &  train\_r2 &  test\_r2 \\
\midrule
         Moody's Investors Service &     0.05869 &    0.33465 &   0.98615 &  0.55041 \\
                     Fitch Ratings &     0.07417 &    0.36611 &   0.97411 &  0.41245 \\
Standard \& Poor's Ratings Services &     0.13550 &    0.27965 &   0.92238 &  0.67990 \\
        Egan-Jones Ratings Company &     0.10015 &    0.26836 &   0.94611 &  0.61277 \\
\bottomrule
\end{tabular}

    \label{tab:Omega$^2$_reg_results}
\end{table}

The regression results demonstrated sub-milli scale root mean squared error (RMSE) and high explanatory power, with R² values above 0.90 on training sets and above 0.60 on test sets across all agencies. This consistency reflects that Omega$^2$ effectively captured nonlinear dependencies between financial ratios and agency-assigned credit scores.

The modest degradation between training and testing R² values indicates realistic generalization rather than overfitting. Importantly, performance trends were highly stable across agencies, proving that Omega$^2$ maintained robustness despite institutional or reporting differences.

We also observed that Moody’s and Standard \& Poor’s datasets produced slightly higher R² scores than Fitch and Egan-Jones. This variation likely results from the richer rating histories and more granular financial disclosures available in the first two agencies. However, overall predictive dispersion remained narrow, reinforcing the framework’s cross-agency consistency.

\subsection{Cross-agency Generalization and Temporal Robustness}
A central objective of Omega$^2$ was to ensure consistent learning behavior across different rating institutions and time horizons. Our temporal group-split strategy guaranteed that no future information leaked into the training window. All evaluations were performed on chronologically unseen data, mimicking real-world forward deployment. The small variance in AUC and RMSE across agencies confirmed that Omega$^2$ generalizes structural financial patterns rather than memorizing agency-specific labels.

This consistency validates that the architecture captures stable, economically meaningful trends that persist across reporting cycles. The approach not only ensures statistical soundness but also aligns with institutional requirements for temporal integrity in credit modeling.

\subsection{Feature-level Interpretation and Financial Insights}
To interpret how Omega$^2$’s predictions align with traditional financial reasoning, we aggregated SHAP-based feature importances across all models and agencies. The unified importance ranking identified operating margin, return on equity, current ratio, and long-term debt to capital as the top four predictors influencing both classification and regression outputs.

This alignment between data-driven insights and economic theory demonstrates that Omega$^2$ internalized the same credit drivers that human analysts rely on—profitability, leverage, and liquidity. The feature-level coherence between classification and regression models ensures interpretability and builds institutional trust.

We further verified the stability of these importances over time. Their rankings remained largely invariant across temporal folds, confirming that Omega$^2$ captures persistent, fundamental credit determinants rather than short-lived correlations.

\subsection{Summary of Results}
Across all experiments, Omega$^2$ achieved high temporal consistency, robust cross-agency reproducibility, and transparent interpretability. The framework’s classification pipeline delivered AUC values above 0.93, while its regression counterpart achieved a mean test $R^2$ above 0.60 and sub-milliscale RMSE. These results collectively establish Omega$^2$ as a unified learning system capable of both binary credit classification and continuous risk quantification within a single reproducible architecture.

All code, data splits, and model configurations used in this study are available in the public GitHub repository, ensuring full transparency and reproducibility. The consistency of outcomes across independent agencies underscores Omega$^2$’s value as a scalable, institution-grade foundation for credit-risk analytics that combine interpretability with predictive strength.

\noindent\textbf{Public GitHub Repository of Omega\textsuperscript{2}:} 
\url{https://github.com/Swarnendu-Bhattacharjee/OmegaSquared}

\section{Conclusions and Future Work}

This study presented Omega\textsuperscript{2}, a unified and temporally consistent framework for corporate credit-risk modelling across multiple rating agencies. 
Unlike conventional credit-scoring architectures that rely solely on static financial snapshots, Omega\textsuperscript{2} was explicitly designed to capture evolving, nonlinear dependencies within firm-level financial indicators through temporally constrained learning. 
The incorporation of gradient-boosting architectures—CatBoost, LightGBM, and XGBoost—ensured both predictive robustness and interpretability across diverse rating ecosystems.

The results demonstrate that when temporal integrity and agency-specific calibration are rigorously maintained, reliable and reproducible forecasting of credit quality can be achieved without the need for overly complex or opaque model structures. 
By validating the system across Moody’s, Standard \& Poor’s, Fitch, and Egan-Jones datasets, the study confirms that the same methodological backbone can generalize effectively across heterogeneous credit-reporting regimes. 
This cross-agency reproducibility represents a core step toward transparent and institution-grade financial intelligence systems.

Beyond predictive accuracy, Omega\textsuperscript{2} establishes a replicable experimental paradigm for evaluating model bias, stability, and fairness under real-world temporal constraints. 
All reported outcomes can be regenerated directly from the publicly accessible repository, ensuring end-to-end reproducibility in accordance with open-science standards. 
The approach thereby shifts focus from numerical optimization alone to methodological soundness—emphasizing transparency, interpretability, and ethical consistency in automated risk assessment.

Future development of Omega\textsuperscript{2} will extend beyond structured financial data to integrate contextual and macroeconomic intelligence layers. 
Incorporating textual analysis of corporate disclosures, regulatory filings, and ESG-linked metrics will enable a more holistic representation of institutional credit health. 
Additional research will focus on quantifying and mitigating temporal and demographic bias, establishing explainability benchmarks, and evolving the system into a self-adaptive analytical engine for dynamic credit-risk governance.

In summary, Omega\textsuperscript{2} provides a reproducible, temporally aware foundation for institutional-grade credit assessment. 
By unifying rigorous quantitative modelling with interpretable, bias-conscious evaluation, the framework paves the way toward a transparent and ethically aligned generation of financial AI systems.

\section*{Bibliography}
Biecek, P., Chlebus, M., Gajda, J., Gosiewska, A., Kozak, A., Ogonowski, D., Sztachelski, J., Wojewnik, P., ``Enabling Machine Learning Algorithms for Credit Scoring - Explainable Artificial Intelligence (XAI) methods for clear understanding complex predictive models'', ArXiv abs/2104.06735 (2021).

Demajo, L. M., Vella, V., Dingli, A., ``Explainable AI for Interpretable Credit Scoring'', Computer Science \& Information Technology (CS \& IT) Credit Scoring and Its Applications, 2020.

Duanyu Feng, Yongfu Dai, Jimin Huang, Yifang Zhang, Qianqian Xie, Weiguang Han, Zhengyu Chen, Alejandro Lopez-Lira, Hao Wang, ``Empowering Many, Biasing a Few: Generalist Credit Scoring through Large Language Models'', arXiv:2310.00566 [cs.LG].

Melsom, B., Vennerod, C. B., de Lange, P. E., Hjelkrem, L. O., ``Explainable artificial intelligence for credit scoring in banking'', Journal of Risk 25(2), 1-25, 2022.

Peter Martey Addo, Dominique Guegan, Bertrand Hassani, ``Credit Risk Analysis Using Machine and Deep Learning Models'', Risks, 6(2), 2018.

Qiu, Z., Li, Y., Ni, P., Li, G., ``Credit Risk Scoring Analysis Based on Machine Learning Models'', 6th International Conference on Information Science and Control Engineering (ICISCE), 2019.

Thomas, L. C., Crook, J., Edelman, D. B., ``Credit Scoring and Its Applications'', Society for Industrial and Applied Mathematics, Philadelphia, Pa., 2002.

Wallis, M., Kumar, K., Gepp, A., ``Credit Rating Forecasting Using Machine Learning Techniques'', In book: Managerial Perspectives on Intelligent Big Data Analytics (pp.180-198), 2019.

\noindent\textbf{Public GitHub Repository of Omega\textsuperscript{2}:} 
\url{https://github.com/Swarnendu-Bhattacharjee/OmegaSquared}

\noindent\textbf{Public Kaggle Dataset:} 
\url{https://www.kaggle.com/datasets/kirtandelwadia/corporate-credit-rating-with-financial-ratios}

\section*{Appendix}
The measures for evaluating accuracy are the overall accuracy (percentage of correct classifications) and Cohen's Kappa statistic (Kappa). Kappa is an accuracy measure that considers both the observed accuracy of the model and the expected accuracy. It is calculated as:

\[
\frac{\text{Observed Accuracy} - \text{Expected Accuracy}}{1 - \text{Expected Accuracy}}
\]

The inclusion of the expected accuracy allows the statistic to adjust to imbalanced data sets that do not have an equal number of data points in each category. This is particularly important given that credit rating forecasting often involves imbalanced data, as mentioned earlier in the Background. Using the following equation, the Kappa statistic considers the expected chance of agreement between the truth and the model:

\[
K = (P_o - P_e)/(1 - P_e)
\]

where $P_o$ represents the observed agreement and $P_e$ represents the expected agreement.

Data pre-processing, the second phase in the pipeline, stands out as one of the most time-consuming aspects, involving several critical sub-tasks: data cleaning, feature extraction, feature selection, feature engineering, and data segregation. Let's delve deeper into each:
1) Data cleaning
2) Feature extraction
3) Feature selection
4) Feature engineering

\begin{figure}[H]
    \centering
    \includegraphics[width=0.7\textwidth]{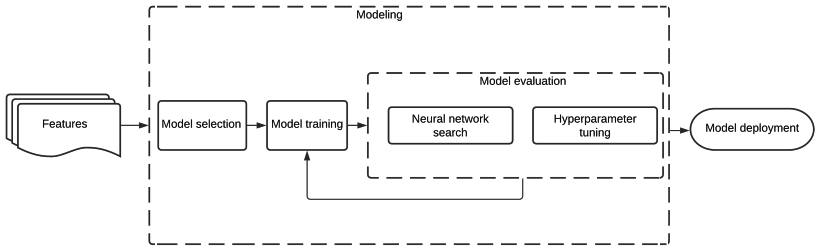}
\end{figure}

During deployment, continuous monitoring of the model's performance in real-world scenarios is crucial. This monitoring allows for adjustments and calibrations to be made as necessary. This process represents an ongoing model cycle, starting from data ingestion and culminating in deployment.

\textbf{Extreme Gradient Boosting (XGBoost):}

\noindent Ø XGBoost stands as a potent and highly efficient gradient-boosting algorithm crafted to address a diverse array of machine learning challenges. Similar to random forest, it serves both regression and classification tasks and has garnered widespread acclaim for its outstanding performance in predictive modeling competitions and practical applications. Particularly adept at handling structured/tabular data, XGBoost is celebrated for its resilience, scalability, and adeptness at capturing intricate patterns.

\noindent Ø How it Functions: XGBoost constructs an ensemble of weak predictive models, typically decision trees, sequentially. Each subsequent model endeavors to rectify errors made by its predecessors. The core tenets of XGBoost are outlined below:

\noindent Ø Gradient Boosting: XGBoost leverages gradient boosting, which iteratively enhances the ensemble by minimizing a loss function.

\noindent Ø Regularization: Incorporating L1 (lasso) and L2 (ridge) regularization terms into the loss function, XGBoost regulates overfitting.

\noindent Ø Feature Importance: XGBoost furnishes insights into feature importance, enabling comprehension of each feature's contribution to model predictions.

\noindent Ø Cross-Validation: XGBoost facilitates k-fold cross-validation for evaluating and optimizing model performance.

\begin{figure}[H]
    \centering
    \includegraphics[width=\textwidth]{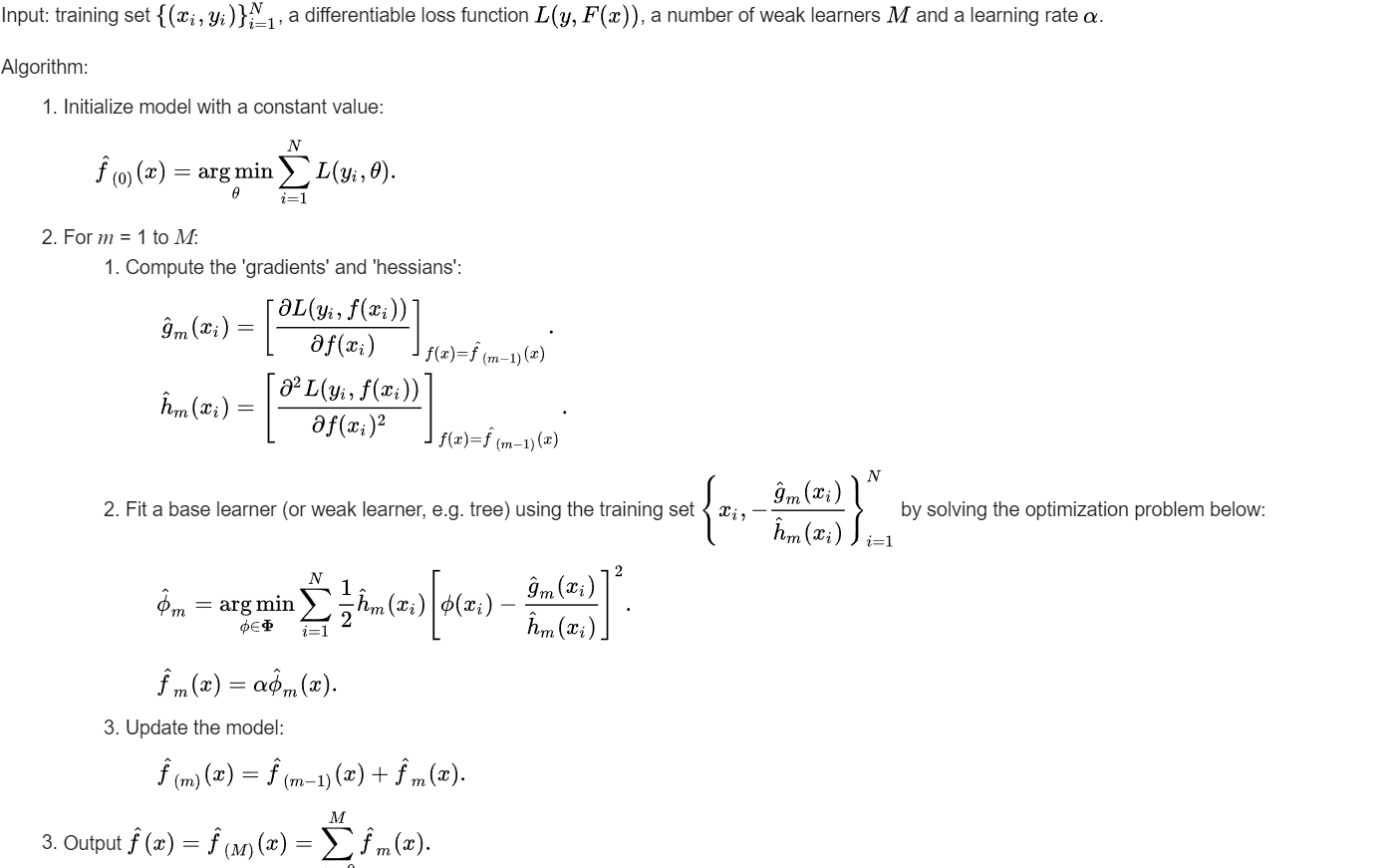}
\end{figure}

\begin{figure}[H]
    \centering
    \includegraphics[width=0.8\textwidth]{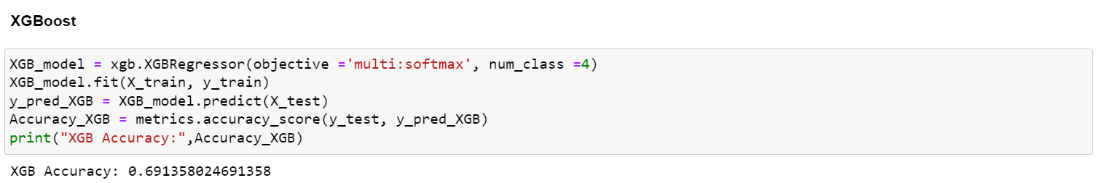}
\end{figure}

\begin{figure}[H]
    \centering
    \includegraphics[width=0.8\textwidth]{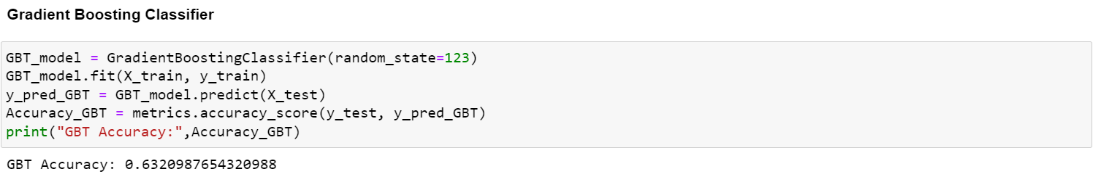}
\end{figure}

\end{document}